\documentclass[10pt,twocolumn,letterpaper]{article}

\usepackage{cvpr}
\usepackage{times}
\usepackage{epsfig}
\usepackage{graphicx}
\usepackage{amsmath}
\usepackage{amssymb}

\usepackage{tabularx}
\usepackage{epsfig}
\usepackage{lscape}
\usepackage{subcaption}

\usepackage{algorithm}
\usepackage{algorithmic}

\usepackage{wrapfig}
\usepackage{fixltx2e}
\graphicspath{{images/}}
\DeclareGraphicsExtensions{.pdf,.png,.jpg,.gif}

\usepackage[pagebackref=true,breaklinks=true,letterpaper=true,colorlinks,bookmarks=false]{hyperref}

 \cvprfinalcopy % *** Uncomment this line for the final submission

\def\cvprPaperID{25} % *** Enter the CVPR Paper ID here

\ifcvprfinal\fi
\begin{document}

%%%%%%%%% TITLE
\title{Parametric Shape Modeling and Skeleton Extraction with Radial Basis Functions using Similarity Domains Network}

\author{Sedat Ozer\\
{\tt\small sedat@csail.mit.edu}}

%\and

\maketitle
%\thispagestyle{empty}

%%%%%%%%% ABSTRACT
\begin{abstract}

We demonstrate the use of similarity domains (SDs) for shape modeling and skeleton extraction. SDs are recently proposed and they can be utilized in a neural network framework to help us analyze shapes. SDs are modeled with radial basis functions with varying shape parameters in Similarity Domains Networks (SDNs). In this paper, we demonstrate how using SDN can first help us model a pixel-based image in terms of SDs and then demonstrate how those learned SDs can be used to extract the skeleton of a shape.

\end{abstract}

%%%%%%%%% BODY TEXT
\section{Introduction}

\begin{figure}[t!]
  \centering
\begin{subfigure}{.5\linewidth}
  \centering
  \includegraphics[width=\linewidth]{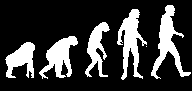}
    \vspace{-0.24in}
   \caption{\small Binary input image}
  \label{fig:sfig1}
\end{subfigure}%
~
\begin{subfigure}{.5\linewidth}
  \centering
  \includegraphics[width=\linewidth]{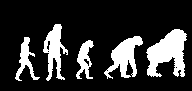}
   \vspace{-0.24in}
  \caption{\small Altered image using SDs}
  \label{fig:sfig2}
\end{subfigure}
~
\begin{subfigure}{.5\linewidth}
  \centering
  \includegraphics[width=\linewidth]{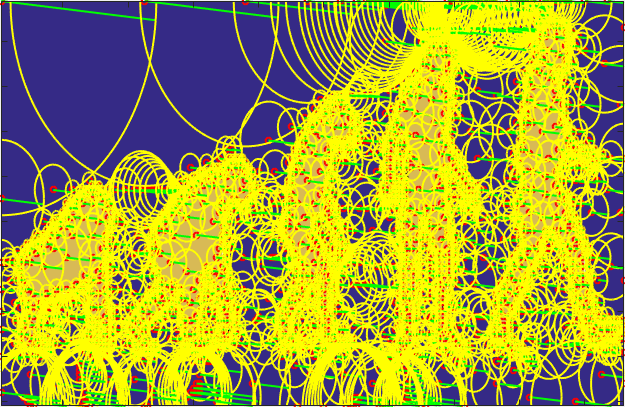}
    \vspace{-0.24in}
  \caption{\small Visualization of all the SDs}
  \label{fig:sfig3}
\end{subfigure}%
~
\begin{subfigure}{.5\linewidth}
  \centering
  \includegraphics[width=\linewidth]{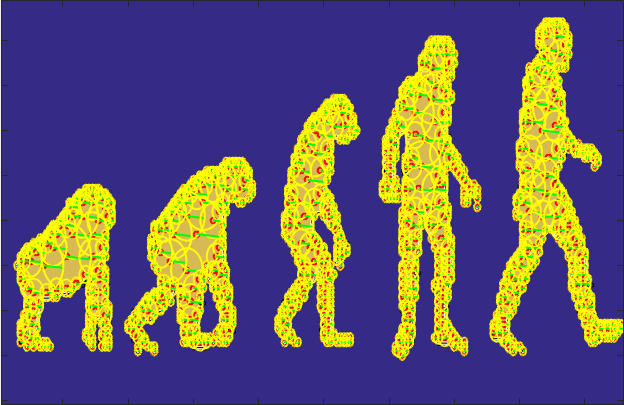}
     \vspace{-0.24in}
    \caption{\small Visualization of the (+) SDs}
  \label{fig:sfig4}
\end{subfigure}
  \vspace{-0.14in}

\caption{The use of the shape parameters of SDN on a binary image is shown. (a) Original binary image. (b) The altered image by utilizing the SDN's shape parameters. Each object is scaled and shifted at different scales. We run a region growing algorithm to group the kernel parameters for each object and then individually scale and shift them. (c) All the computed shape parameters of the input image are visualized. (d) Visualization of the foreground parameters.}
\label{fig:Usingsimilaritydomains}
\end{figure}

Recent advances in deep learning moved attention to the neural networks based solutions for shape understanding, shape analysis and parametric shape modeling. Radial basis networks (RBNs) are a particular set of neural networks using radial basis function (RBF) kernels and in this paper, we introduce a novel shape modeling algorithm based on RBNs. RBFs have been used in the literature for many classification tasks including the original LeNET architecture \cite{lecun1998gradient}. While RBFs are useful in modeling surfaces and classification tasks as in \cite{ozer2010supervised,jiang2018parametric, yoo2015optimized, botsch2005real,ozer2009prostate, ozer2007classification}, there are many challenges associated with utilizing RBFs in neural networks for parametric shape modeling. Two of those challenges include: (I) estimating the optimal number of RBFs (e.g., the number of circles in our figures) to be used in the network along with their optimal center values, and (II) estimating the optimal RBF kernel parameters by relating them to shapes geometrically. The kernel parameters are typically known as the scale or the shape parameter (representing the radius of a circle in this paper) and used interchangeably in the literature. The standard RBNs as defined in \cite{lippmann1989pattern} applies the same kernel parameter to each and all basis functions used in the architecture. Recent literature focused on using multiple kernels with their own kernel parameters as in \cite{fu2010sparse} and \cite{bach2004multiple}. While the idea of utilizing different kernels with different parameters has been heavily studied in the literature under the "Multiple Kernel Learning" (MKL) framework as formally modeled in \cite{bach2004multiple}, there are not many efficient approaches and available implementations focusing on utilizing multiple kernels with their own parameters in RBNs for shape modeling. Recently, the work in \cite{ozer2019similarity} combined the optimization advances achieved in the kernel machines domain with the radial basis networks and introduced a novel algorithm for shape analysis. In this paper, we call that algorithm as "Similarity Domains Network" (SDN) and discuss its benefits from both shape analysis (see Figure~\ref{fig:Usingsimilaritydomains}) and skeleton extraction perspectives. As we demonstrate in this paper, the computed SDs of SDN can be used to obtain both parametric models for shapes via its SDs and their skeletons without requiring large training samples.

  \vspace{-0.14in}
\section{Related Work}

In this paper, we propose using SDs for both parametric shape modeling and for extracting the skeleton. Our proposed algorithm: SDN is related to both RBNs and kernel machines. Skeleton extraction has been widely studied in the literature as in \cite{cornea2007curve, sundar2003skeleton,saha2016survey,SkelNetOn19}. However, in this paper, we mainly discuss and present our novel algorithm from the RBNs perspective. In the past, the RBN related research mostly focused on computing the optimal single kernel parameter (i.e., the scale or shape parameter) to be used in all of the RBFs used in the network as in \cite{mongillo2011choosing, biazar2017interval}. While the parameter computation for multiple kernels have been heavily studied under the MKL framework in the literature (for examples, see the survey papers: \cite{bucak2014multiple, gonen2011multiple}), the computation of multiple kernel parameters in RBNs has been mostly studied under two main approaches: using optimization or using heuristic methods. For example, in \cite{benoudjit2002width}, the authors proposed using multiple scales as opposed to using a single scale value in RBNs. Their approach utilizes first computing the standard deviation of each cluster (after applying a k-means like clustering on the data) and then using a scaled version of those standard deviations of each cluster as the shape parameter for each  RBF in the network. The work in \cite{bataineh2017neural} also used a similar approach by using the root-mean-square-deviation (RMSD) value between the RBF centers and the data value for each RBF in the network. The authors used a modified orthogonal least squares (OLS) algorithm to select the RBF centers. The work in \cite{fu2010sparse} used k-means algorithm on the training data to choose k centers and used those centers as RBF centers. Then it used separate optimizations for computing the kernel parameters and the kernel weights (see next chapter for the formal definitions). Using additional optimization steps for different set of parameters is costly and makes it harder to interpret those parameters and to relate them to shapes geometrically and accurately. As an alternative solution, the work in \cite{ozer2019similarity} proposed a geometric approach by using the distance between the data samples as a geometric constraint. In \cite{ozer2019similarity}, the author did not use the well known MKL model. Instead, he defined interpretable similarity domains concept using RBFs and developed his own optimization approach with geometric constrains similar to the original Sequential Minimal Optimization (SMO) algorithm \cite{platt1999fast}. Consequently, the SDN algorithm combines both RBN and kernel machine concepts to develop a novel algorithm with geometrically interpretable kernel parameters. In this paper, we propose using SDN for parametric shape modeling and skeleton extraction. Unlike the existing work, instead of applying an initial k-means algorithm or OLS algorithm to compute the kernel centers separately or using multiple cost functions, SDN chooses the RBF centers and their numbers automatically via its sparse modeling and uses a single cost function to be optimized with its geometric constraint. That is where SDN differs from other similar RBN works as they would have issues on computing all those parameters within a single optimization step while automatically adjusting the number of RBFs used in the network sparsely.

\begin{figure}[t]
\centering
\includegraphics[scale=0.23]{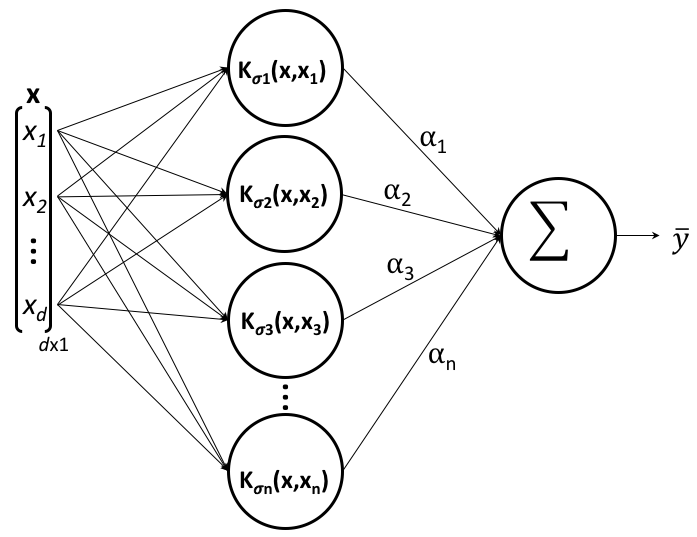}
  \vspace{-0.14in}
\caption{\small An illustration of SDN as a radial basis network. The network contains a single hidden layer. The input layer ($d$ dimensional input vector) is connected to $n$ radial basis functions. The output is the weighted sum of the radial basis functions' outputs.}
\label{fig:SDN}
\end{figure}

\begin{figure*}[t]
\centering
  \begin{subfigure}{.25\linewidth}
  \centering
  \includegraphics[width=\linewidth]{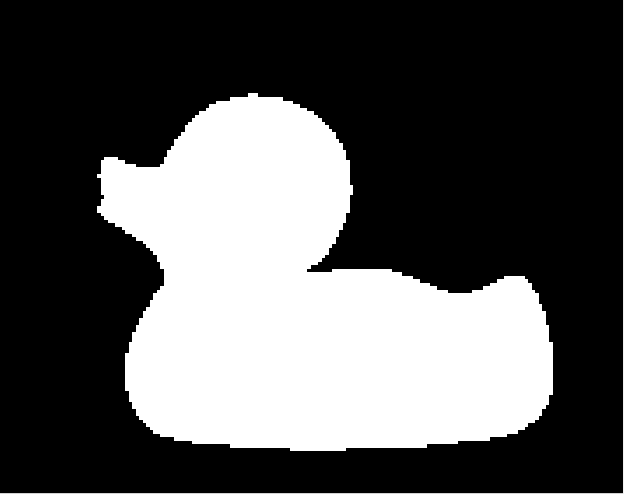}
  \vspace{-0.24in}
  \caption{The input image}
  \label{fig:DuckOriginal}
\end{subfigure}
~
\begin{subfigure}{.33\linewidth}
  \centering
  \includegraphics[width=\linewidth]{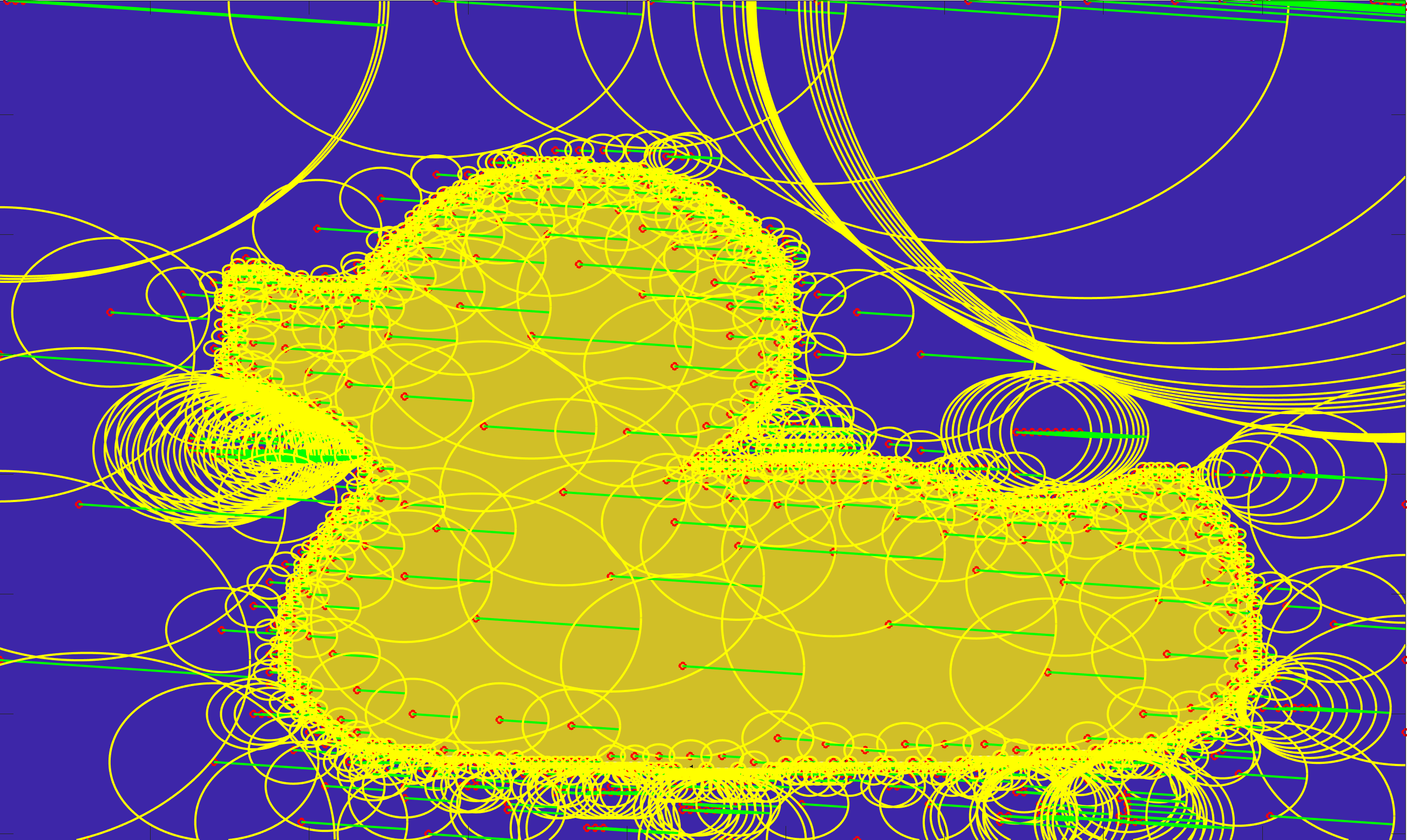}
    \vspace{-0.24in}
  \caption{All $r_i$}
  \label{fig:DuckAllParams}
\end{subfigure}
~
\begin{subfigure}{.33\linewidth}
  \centering
  \includegraphics[width=\linewidth]{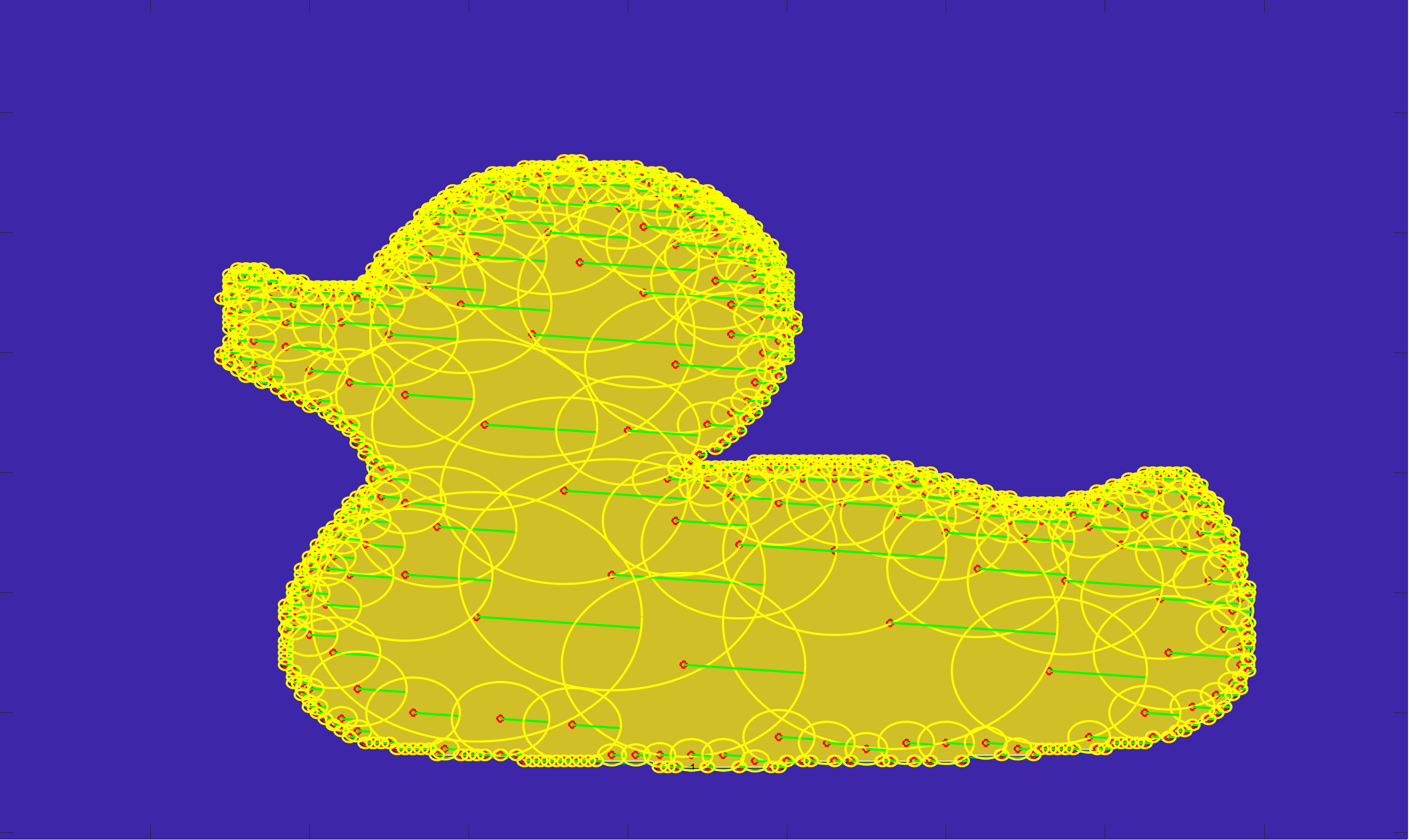}
    \vspace{-0.24in}
  \caption{Foreground $r_i$}
  \label{fig:DuckForegroundParams}
\end{subfigure}%
\vspace{-0.1in}
\caption{\small Visualization of the SDM kernel parameters at $T$= 0.05 with zero pixel error learning. The blue area represents the background and the yellow area represents the foreground. The red dots are the RBF centers and yellow circles around them show the boundaries of SDs. The green lines are the radiuses ($r_i$) of SDs. The $r_i$ are obtained from the computed $\sigma_i$. \textbf{(a)} Original image: 141x178 pixels. \textbf{(b)} Visualization of all the $r_i$ from both background and foreground with total of 1393 centers. \textbf{(c)} Visualization of only the $r_i$ for the object with total of 629 foreground centers (i.e., by using only the 2.51\% of all image pixels). All images are resized to fit into the figure.}
\label{fig:Duck}
\end{figure*}
\vspace{-0.1in}

\section{Similarity Domains Network}

RBNs typically include a single hidden layer using radial basis functions as activation functions and the hidden layer uses $n$ different RBFs. The illustration of SDN as a radial basis network is given in Figure \ref{fig:SDN}. In the figure, the hidden layer uses all of the $n$ training data as an RBF center and then through the sparse optimization, it selects a subset of the training data (e.g., subset of pixels for shape modeling). SDN represents the decision boundary as a weighted combination of Similarity Domains (SDs). A Similarity Domain is a $d$ dimensional sphere in the $d$ dimensional feature space. Each similarity domain is centered at an RBF center and modeled with a Gaussian RBF in SDN. SDN estimates the label $y$ of a given input vector \textbf{x} as $\overline{y}$ as shown below:
  \vspace{-0.1in}
 \begin{equation}
 \label{eq:SVMDecision}
\overline{y}=sign(f(\mathbf{x}))  \text{     and    }  f(\mathbf{x})= \sum\limits_{i=1}^k \alpha_{i}y_{i}K_{\sigma i}(\mathbf{x},\mathbf{x_{i}}),
\end{equation}
where the scalar $\alpha_i$ is a nonzero weight fo the RBF center ${\bf {x}_{i}}$, $y_i \epsilon \{-1,+1\}$ the class label of the training data and $k$ the total number of RBF centers. $K$(.) is the Gaussian RBF kernel defined as:
  \vspace{-0.1in}
\begin{equation}
\label{eq:GaussianKernel}
 \begin{aligned}
 %& g(\mathbf{x}-\mathbf{x_{i}})=K_{\sigma i}(\mathbf{x},\mathbf{x_{i}}) = exp(-\parallel\mathbf{x}-\mathbf{x_{i}}   \parallel^2/\sigma_{i})  \\
  & K_{\sigma i}(\mathbf{x},\mathbf{x_{i}}) = \exp(-\parallel\mathbf{x}-\mathbf{x_{i}}   \parallel^2/\sigma^2_{i})  \\
  \end{aligned}
  \end{equation}
where $\sigma_i$ is the shape parameter for the center ${\bf{x}_{i}} $. The centers are automatically selected among the training data during the training via the following cost function:
  \vspace{-0.1in}
\begin{equation}
\label{eq:SDM}
 \begin{aligned}
 & \underset{\mathbf{\alpha}}{\text{max}}  \text{   }   
 Q(\mathbf{\alpha}) = \sum\limits_{i=1}^n \alpha_i -\frac{1}{2} \sum\limits_{i=1}^n \sum\limits_{j=1}^n  \alpha_i \alpha_j y_i y_j K_{\sigma ij} (\mathbf{x_{i}},\mathbf{x_{j}}), \\
 & \text{subject to:} \text{   }
    \sum\limits_{i=1}^n \alpha_i y_i = 0,  \text{   }   \text{   }
  \text{   }  C \ge \alpha_i \ge 0  \text{ for  } {i}=1,2,...,n, \\
 &  \text{  and } K_{\sigma ij} (\mathbf{x_{i}},\mathbf{x_{j}}) < T, \text{  if } y_iy_j=-1, \text{    } \forall i,j 
    \end{aligned}
  \end{equation}
where $T$ is a constant value assuring that the RBF function yields a smaller value for any given pair of samples from different classes.  The shape parameter $\sigma{ij}$ is defined as  $\sigma{ij}=min(\sigma_i,\sigma_j)$. Further details on SDs and SDN formulation can be found in \cite{ozer2019similarity}.

\section{Parametric Shape Modeling with SDN}
\label{section:ShapeModeling}
The Gaussian RBFs and their shape parameters can be used for parametric modeling of the shapes. For that, we can save and use only the foreground (the shape's) centers and their shape parameters to obtain a one class classifier. The computed centers of SDN can be grouped as $C_1 =  \bigcup\limits_{i=1, y_i \in {+1}}^{s_1} \mathbf{x_{i}}$ and  $C_2 =  \bigcup\limits_{i=1, y_i \in {-1}}^{s_2} \mathbf{x_{i}}$, where $s_1 + s_2 =k$, $s_1$ is the total number of centers from the (+1) class and $s_2$ is the total number of centers from the (-1) class. Since the Gaussian kernel functions now represent local SDs geometrically, the original decision function $f(\mathbf{x})$ can now be approximated by using only $C_1$ (or by using only $C_2$). Therefore, we define the one-class approximation by using only the centers and their associated kernel parameters from the $C_1$ for any given $\mathbf{x}$ as follows:
  \vspace{-0.14in}
 \begin{equation}
 \label{eq:oneclassdecision}
 \begin{aligned}
 &\overline{y} = +1,\text{    }if \text{    } \parallel \mathbf{x}- \mathbf{x_{i}}   \parallel  < \sqrt{a\sigma^2_i} 
  \text{  } ,  \exists \mathbf{x_{i}}  \in S_1 \\
 & otherwise \text{    } \overline{y} = -1,\text{    } \\
  \end{aligned}
  \end{equation}  
  where the SD radius for the $i^{th}$ center $\mathbf{x_{i}}$ is defined as $\sqrt{a\sigma^2_i}$ and $a$ is a domain specific constant. 
  One class approximation examples are given in Figure~\ref{fig:sfig2} where we used only the SDs from the foreground to reconstruct the altered image.

\begin{figure*}[t]

\begin{subfigure}{.19\linewidth}
  \centering
  \includegraphics[width=\linewidth]{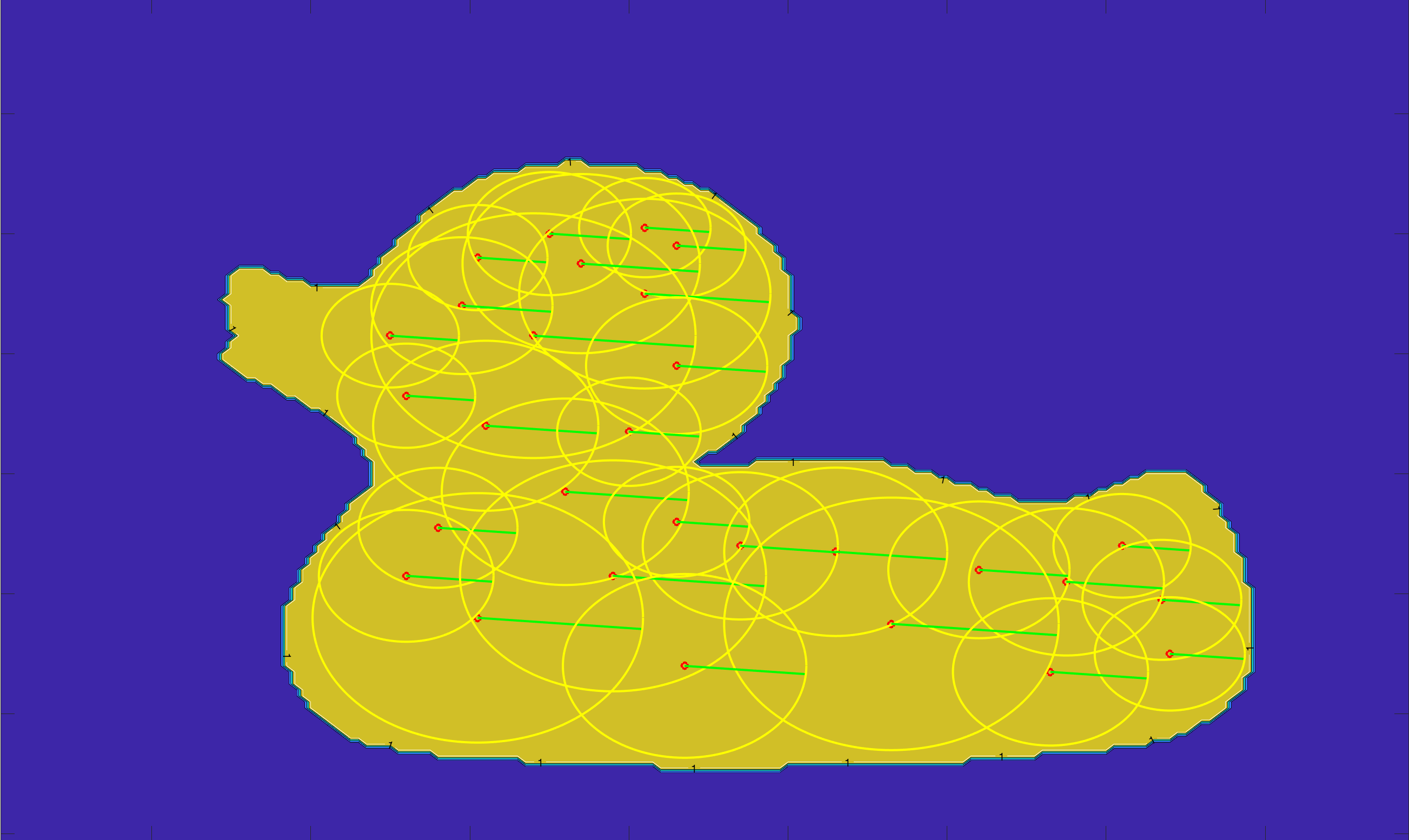}
    \vspace{-0.24in}
  \caption{\small $\sigma^2_i>29.12$}
  \label{fig:DuckT2}
\end{subfigure}
~
\begin{subfigure}{.19\linewidth}
  \centering
  \includegraphics[width=\linewidth]{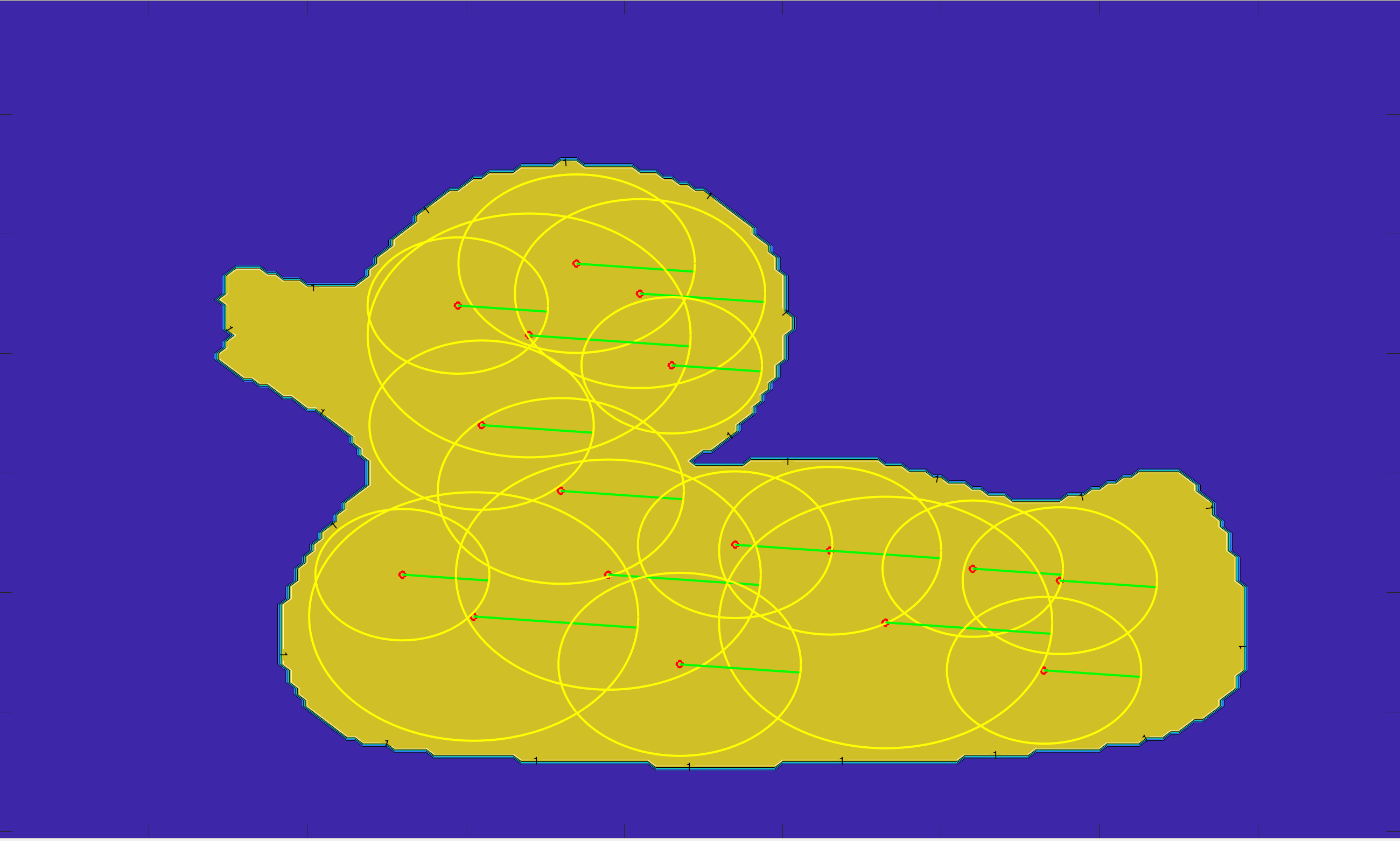}
    \vspace{-0.24in}
  \caption{\small $\sigma^2_i>48.32$}
  \label{fig:DuckT3}
\end{subfigure}%
~
  \begin{subfigure}{.19\linewidth}
  \centering
  \includegraphics[width=\linewidth]{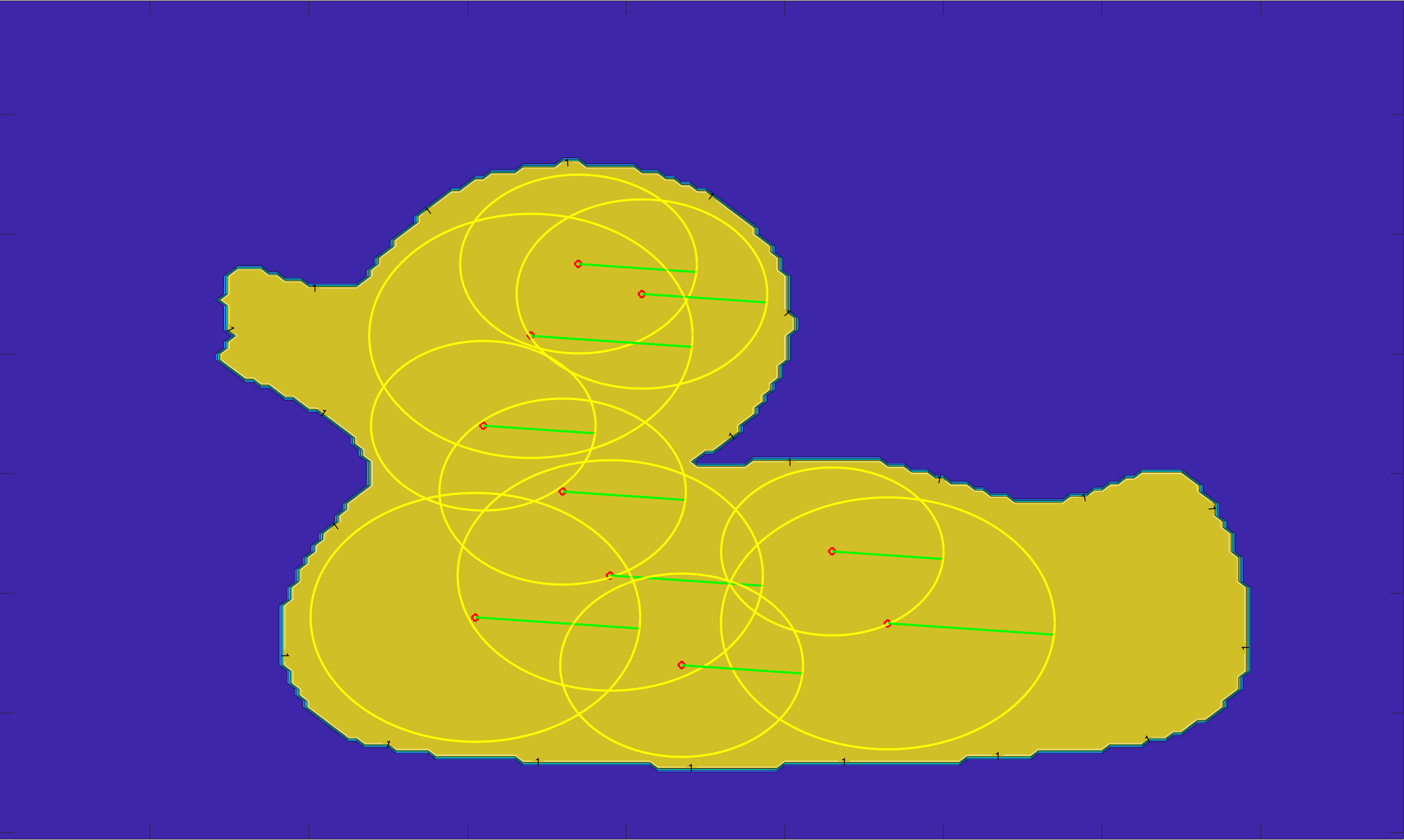}
  \vspace{-0.24in}
  \caption{\small $\sigma^2_i>67.51$}
  \label{fig:DuckT4}
\end{subfigure}
~
\begin{subfigure}{.19\linewidth}
  \centering
  \includegraphics[width=\linewidth]{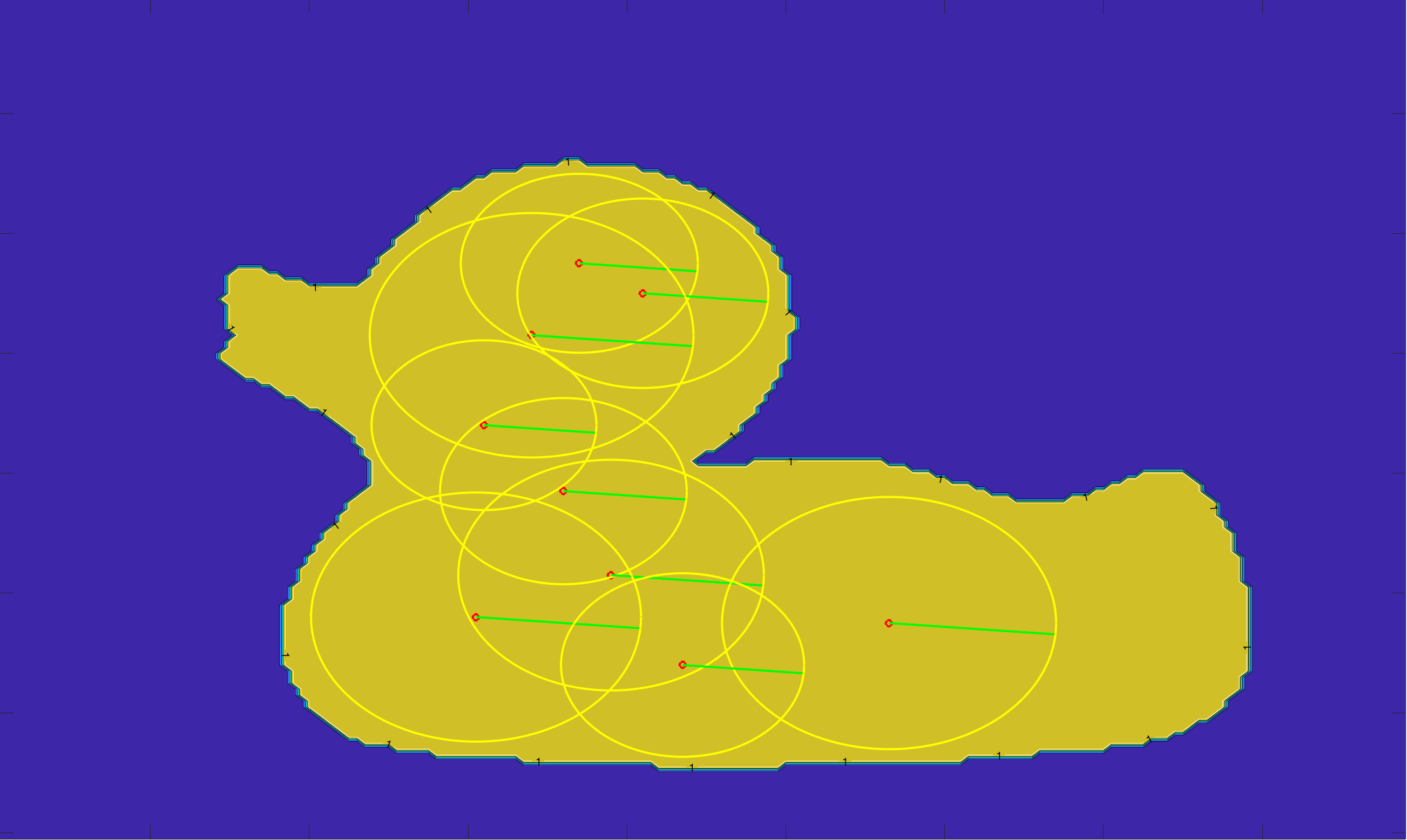}
    \vspace{-0.24in}
  \caption{\small $\sigma^2_i>86.71$}
  \label{fig:DuckT5}
\end{subfigure}
~
\begin{subfigure}{.19\linewidth}
  \centering
  \includegraphics[width=\linewidth]{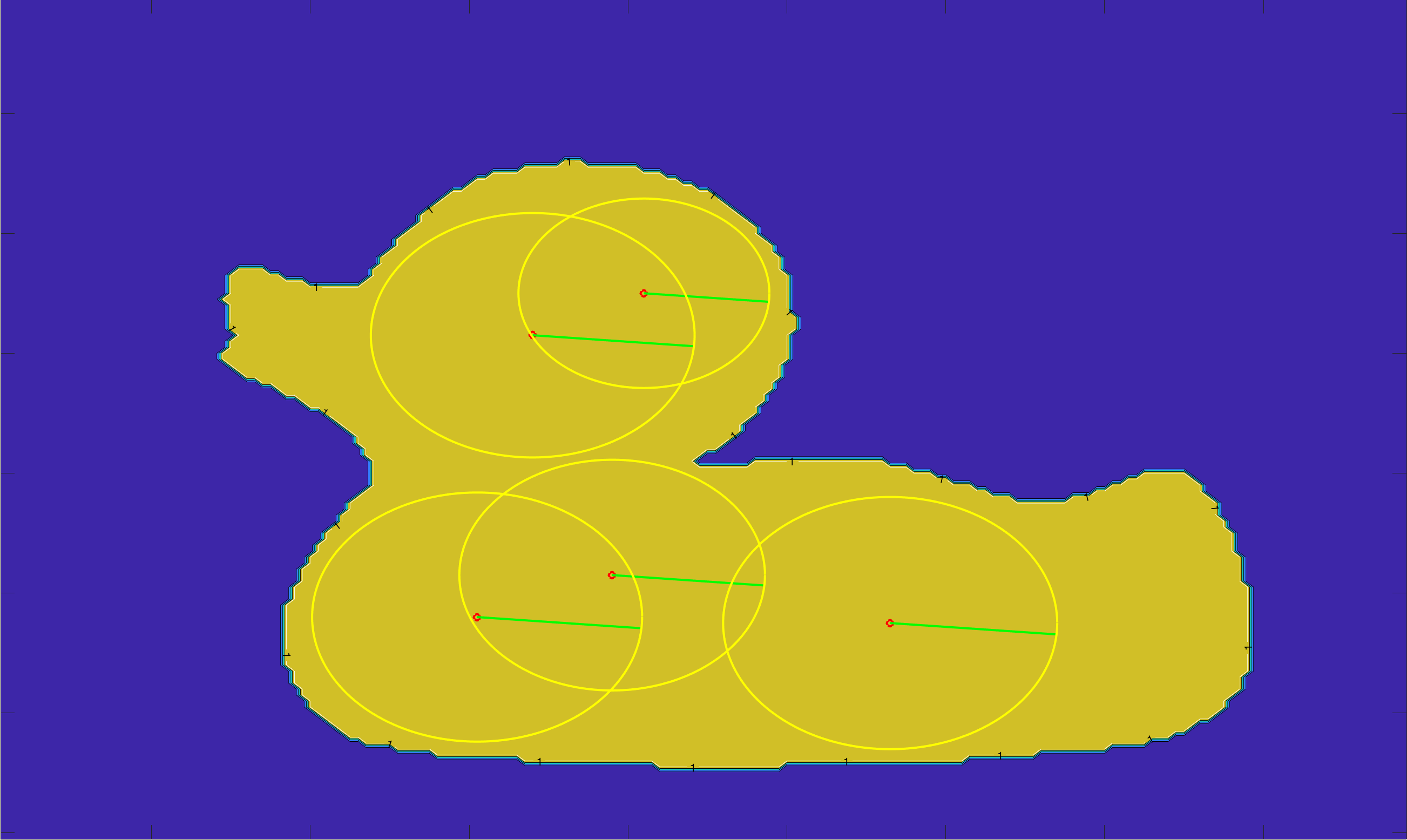}
    \vspace{-0.24in}
  \caption{\small $\sigma^2_i>105.90$}
  \label{fig:DuckT6}
\end{subfigure}%
\\
  \begin{subfigure}{.19\linewidth}
  \centering
  \includegraphics[width=\linewidth]{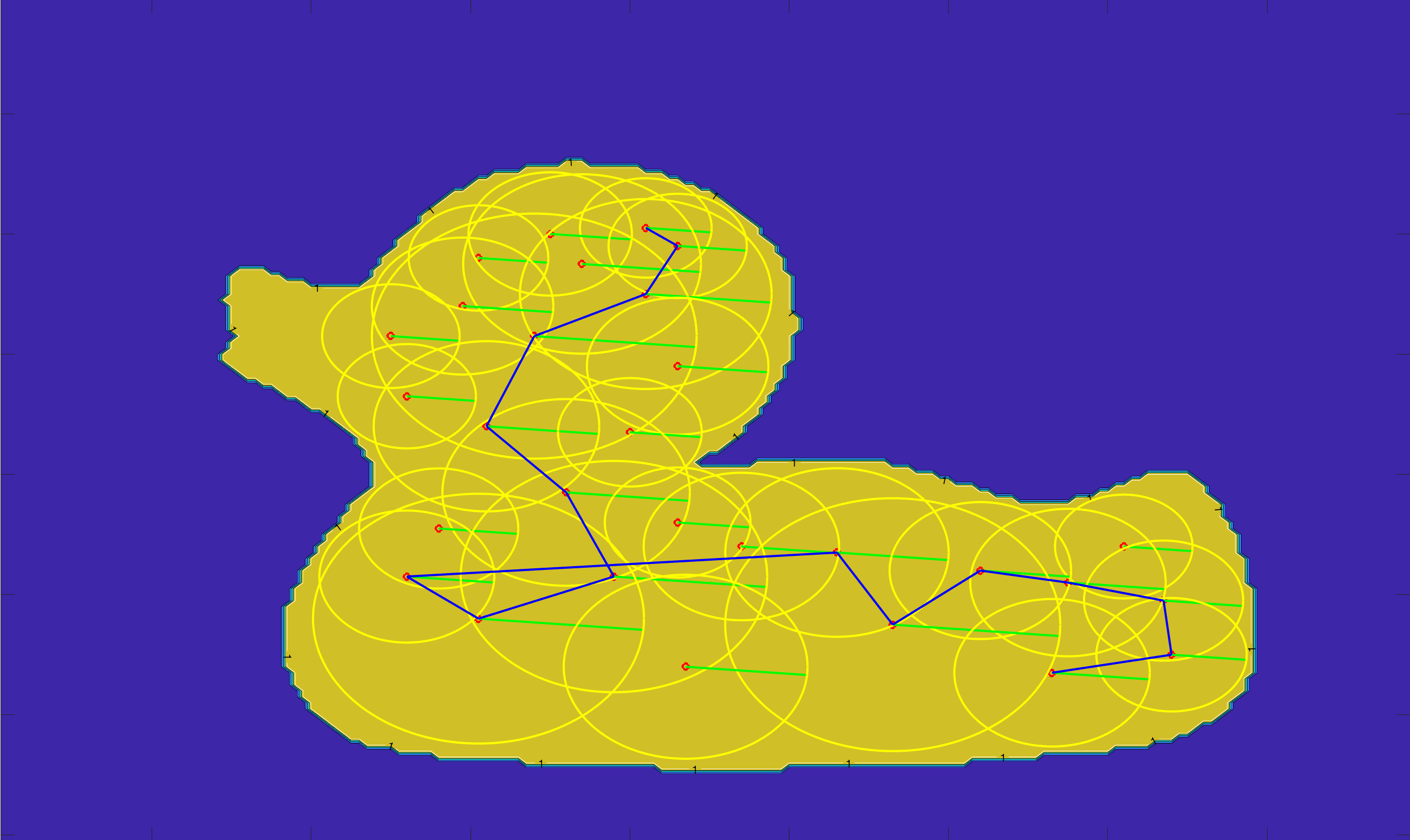}
  \vspace{-0.24in}
  \caption{\small Skeleton for Fig. \ref{fig:DuckT2}}
  \label{fig:DuckT2Skeleton}
\end{subfigure}
~
  \begin{subfigure}{.19\linewidth}
  \centering
  \includegraphics[width=\linewidth]{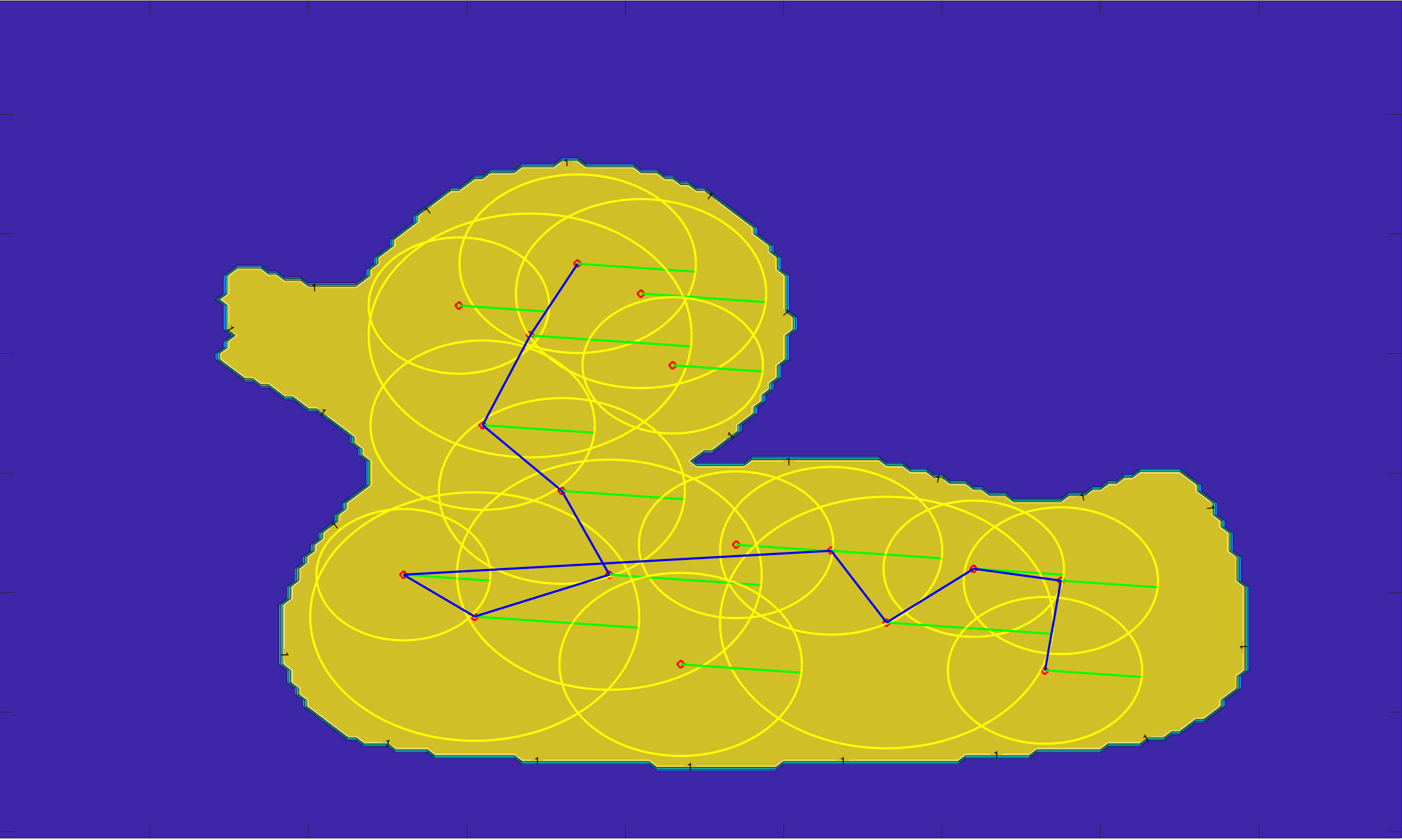}
  \vspace{-0.24in}
  \caption{\small Skeleton for Fig. \ref{fig:DuckT3}}
  \label{fig:DuckT3Skeleton}
\end{subfigure}
~
  \begin{subfigure}{.19\linewidth}
  \centering
  \includegraphics[width=\linewidth]{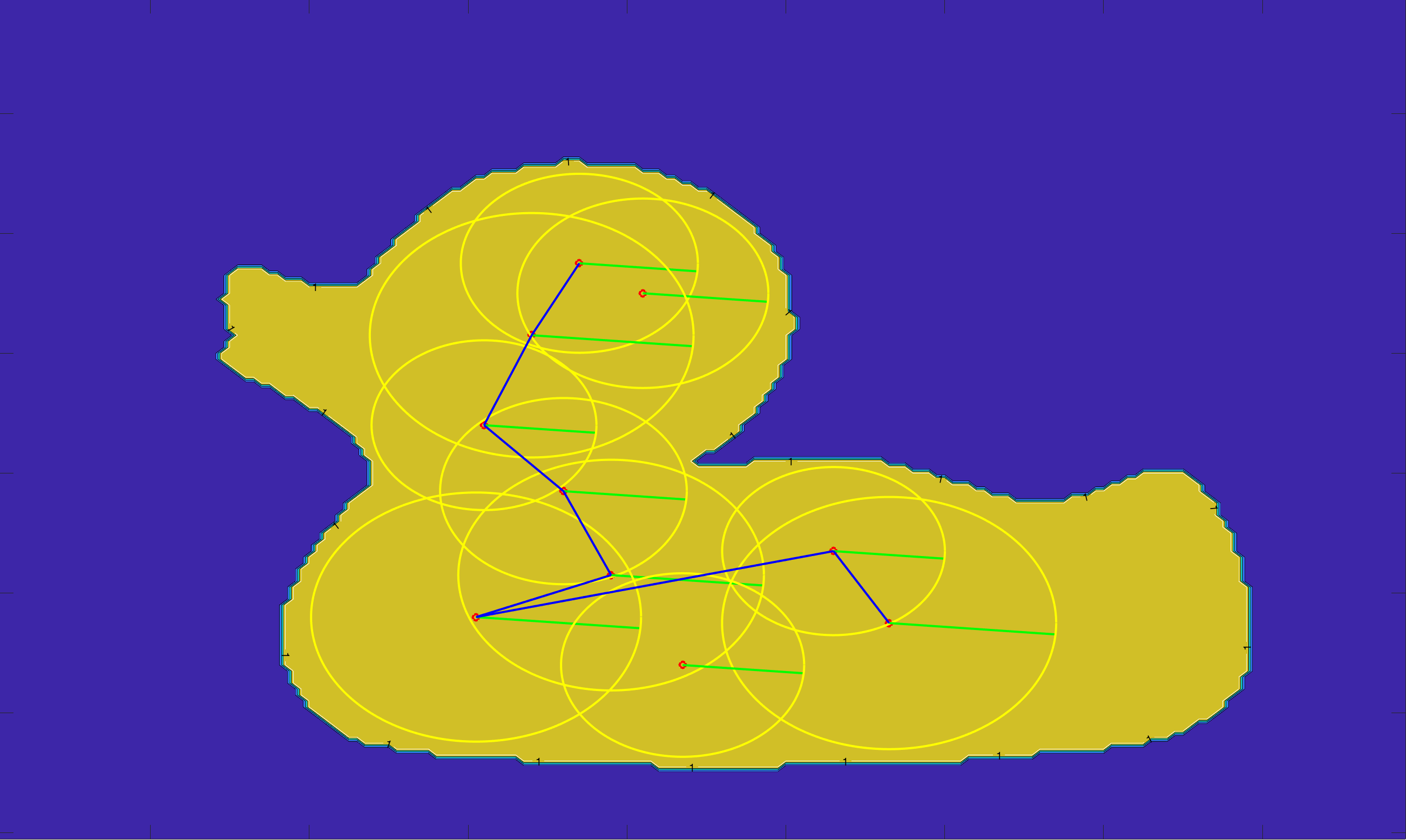}
  \vspace{-0.24in}
  \caption{\small Skeleton for Fig. \ref{fig:DuckT4}}
  \label{fig:DuckT4Skeleton}
\end{subfigure}
~
  \begin{subfigure}{.19\linewidth}
  \centering
  \includegraphics[width=\linewidth]{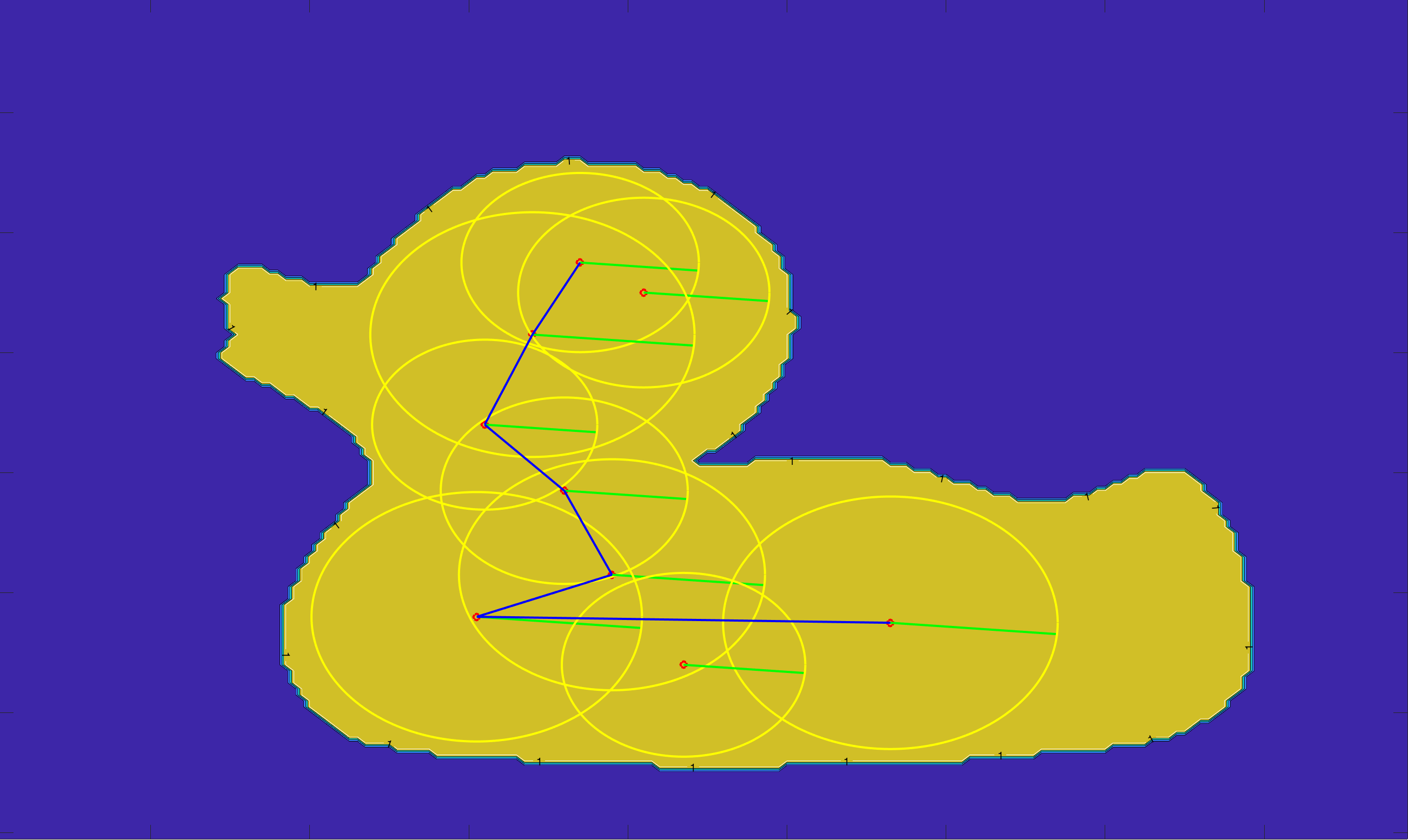}
  \vspace{-0.24in}
  \caption{\small Skeleton for Fig. \ref{fig:DuckT5}}
  \label{fig:DuckT5Skeleton}
\end{subfigure}
~
\begin{subfigure}{.19\linewidth}
  \centering
  \includegraphics[width=\linewidth]{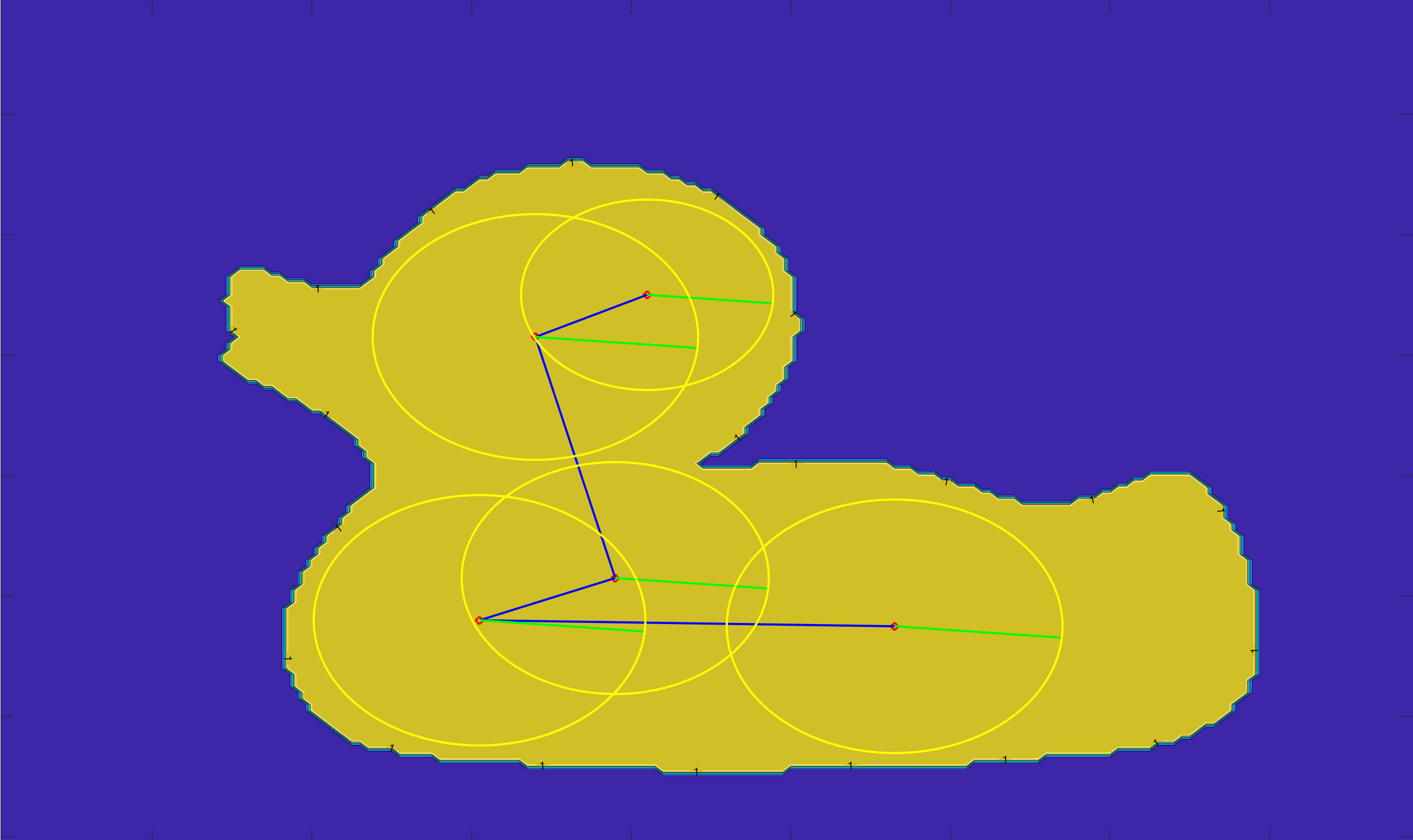}
    \vspace{-0.24in}
  \caption{\small Skeleton for Fig. \ref{fig:DuckT6}}
  \label{fig:DuckT6SkeletonwithSDs}
\end{subfigure}
\vspace{-0.1in}
\caption{\small Visualization of the shape parameters (shown in the first row) after being quantized and thresholded at various values for the image shown in Figure~\ref{fig:DuckOriginal} and their computed skeletons (shown in the second row). Visualization of the shape parameters: (a) for $\sigma^2_i>29.12$; (b) for $\sigma^2_i>48.32$; (c) for $\sigma^2_i>67.51$; (d) for $\sigma^2_i>86.71$; (e) for $\sigma^2_i>105.90$. Extracted skeletons (shown as a blue line) at each of those threshold values are visualized below each image. }
\label{fig:DuckSDs}
\end{figure*}
%------------------------------------------------------------------------

\section{Extracting the Skeleton from SDs}
\label{SkeletonExtraction}
Once learned and computed by the SDN, the Similarity Domains (SDs) can be used to obtain a representation of a shape's skeleton. For that purpose, we first bin the computed shape parameters ($\sigma^2_i$) into $m$ bins (in our experiments $m$ is set to 10). Since typically the majority of the similarity domains lay around the object (or shape) boundary, they appear in small values. Eliminating them at first, gives us a lesser number of SDs to consider for skeleton extraction. After eliminating those small SDs and their computed parameters with a simple thresholding process, we connect the centers of the remaining SDs by tracing the overlapping SDs. In the case of remaining non-overlapping SDs, we connect the closest SDs.

\section{Experiments}
Here, we demonstrate how to use SDN for parametric shape learning from a given single input image. Since it is hard to model shapes with the standard RBNs, and since there is no good RBN implementation was available to us, we did not use any RBN network in our experiments. The standard RBNs (as discussed earlier) have many issues and many individual steps to compute the RBN parameters including the total number of RBF centers and finding the center values along with the computation of the shape parameters at those centers. However, comparison of kernel machines (SVM) and SDN on shape modeling was already studied in the literature before (see \cite{ozer2019similarity}). Therefore, in this section, we focus on parametric shape modeling and skeleton extraction from SDs by using SDNs. All the images are resized to fit into the figures.

\subsection{Parametric Shape Modeling with SDs} 
We first demonstrate visualizing the computed shape parameters of SDN on a sample image in Figure \ref{fig:Duck}. Figure \ref{fig:DuckOriginal} shows the original input image. We used each image pixel's 2D coordinate as the training input, and its color (being black or white) as the training labels. SDN is trained at T=0.05. SDN learned and modeled the shape and reconstructed it with zero pixel error by using 1393 SDs. Pixel error is the total number of wrongly classified pixels in the image. Figure \ref{fig:DuckAllParams} visualizes all the computed shape parameters of the RBF centers of SDN as circles and Figure \ref{fig:DuckForegroundParams} visualizes the ones for the foreground only. The radius of a circle in all figures is computed as $\sqrt{a\sigma^2_i}$ where $a=2.85$. We found the value of $a$ through a heuristic search and noticed that 2.85 suffices for all the shape experiments that we had. There are total of 629 foreground RBF centers computed by SDN (only 2.51\% of all the input image pixels).

\subsection{Skeleton Extraction From the SDs} 
\begin{figure}[t]
  \begin{subfigure}{.305\linewidth}
  \centering
  \includegraphics[width=\linewidth]{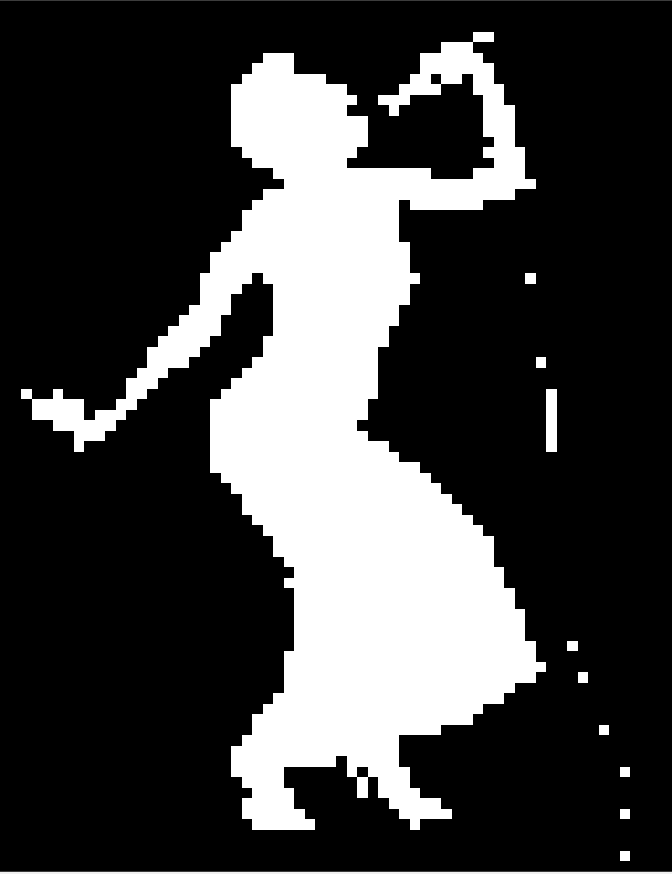}
  \vspace{-0.24in}
  \caption{\small Input Image}
  \label{fig:SingingLady}
\end{subfigure}
~
\begin{subfigure}{.31\linewidth}
  \centering
  \includegraphics[width=\linewidth]{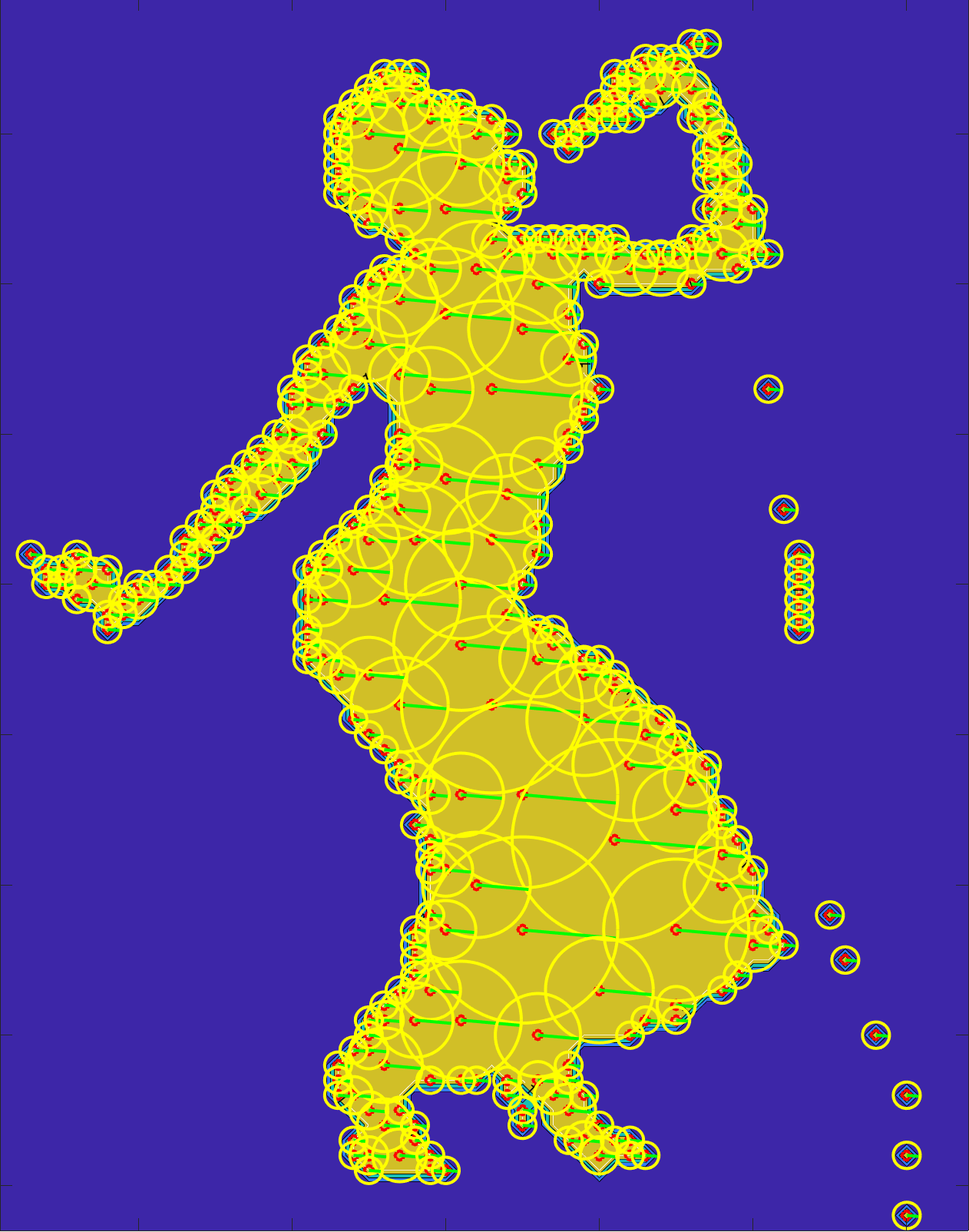}
    \vspace{-0.24in}
  \caption{\small   $\sigma^2_i > 0 $ }
  \label{fig:SingingLadyT3}
\end{subfigure}
~
\begin{subfigure}{.31\linewidth}
  \centering
  \includegraphics[width=\linewidth]{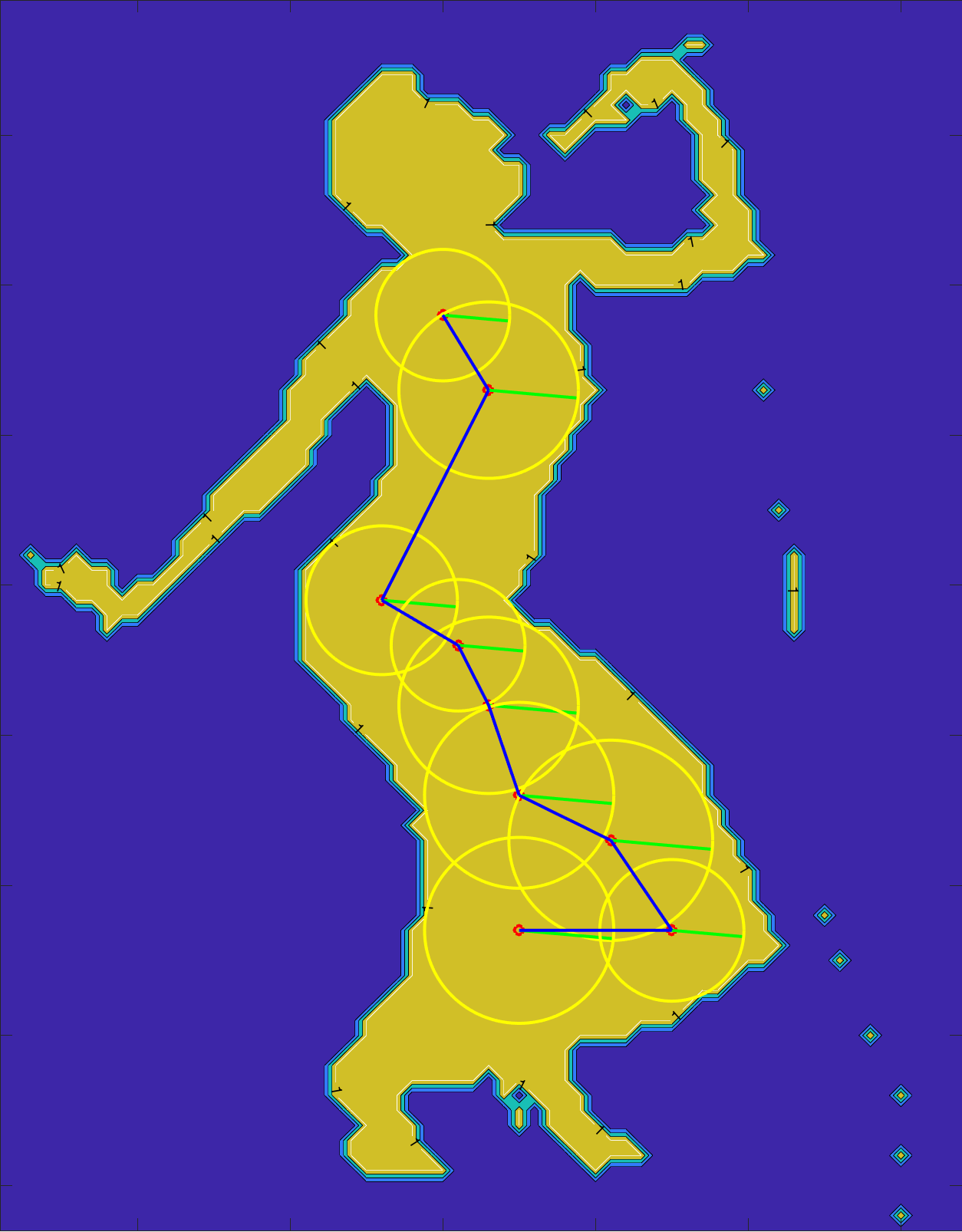}
    \vspace{-0.24in}
  \caption{\small  for $\sigma^2_i > 6.99$ }
  \label{fig:SingingLadyT4}
\end{subfigure}

\vspace{-0.1in}
\caption{\small Visualization of the skeleton (shown as blue line) extracted from SDs on another image. (a) Input image: 64 x 83 pixels. (b) Foreground SDs. (c) Skeleton for $\sigma^2_i > 6.99$. }
\label{fig:SingingLadySkeleton}
\end{figure}

Next, we demonstrate the skeleton extraction from the computed similarity domains as a proof of concept. Extracting the skeleton from the SDs as opposed to extracting it from the pixels, simplifies the computations as SDs are only a small portion of the total number of pixels (reducing the search space). To extract the skeleton from the computed SDs, we first quantize the shape parameters of the object into 10 bins and then starting from the largest bin, we select the most useful bin value to threshold the shape parameters. The remaining SD centers are connected based on their overlapping similarity domains. If multiple SDs overlap inside the same SD, we look at their centers and we ignore the SDs whose centers fall within the same SD (accepted the original SD center). That is why some points are not considered as a part of the skeleton in Figure~\ref{fig:DuckSDs}. First row in Figure~\ref{fig:DuckSDs} demonstrates the remaining SD centers and their radiuses at various thresholds. The second row in the figure visualizes the extracted skeletons (shown as a blue line) from the SDs as explained in Section \ref{SkeletonExtraction}. Another example is shown in Figure~\ref{fig:SingingLadySkeleton}. The learned SDs are thresholded and the corresponding skeleton as extracted from the remaining SDs are visualized as a blue line.

\begin{table}[ht]
\caption{\small Bin centers for the quantized foreground shape parameters ($\sigma^2_i$) and the total number of shape parameters that fall in each bin for the image in Fig.~\ref{fig:DuckOriginal}.}
\vspace{-0.1in}
\centering
\resizebox{\columnwidth}{!}{
\begin{tabular}{l c c c c c c c c c c}
\hline\hline
Bin Center:                  &      9.93    &  29.12    & 48.32     & 67.51   & 86.71  & 105.90   &  125.09  &  144.29   & 163.48  & 182.68\\ 
\hline
Total Counts:               &       591    & 18         & 7       & 3          & 2       & 4         &  0           & 0          & 1         & 3  \\

\hline
\end{tabular}
}
\label{table:filtering}
\end{table}

\vspace{-0.12in}

%\vspace{-0.05in}
\section{Conclusion}

In this paper, we introduced how the computed SDs of the SDN algorithm can be used to extract skeleton from shapes for the first time as a proof of concept. Instead of using and processing all the pixels to extract the skeleton of a shape, we propose to use SDs (a subset of the pixels) to extract the skeleton. The RBF shape parameters of SDN are used to define SDs and they can be used to model a shape as described in Section \ref{section:ShapeModeling} and as visualized in our experiments. While the presented skeleton extraction algorithm is a naive solution to demonstrate the use of SDs, future work will focus on presenting more elegant solutions to extract the skeleton from SDs. SDN is a novel classification algorithm and has potential in many shape analysis applications besides the skeleton extraction. A shape can be modeled parametrically by using SDNs via shape parameters and RBF centers. A further reduction in parameters can be obtained with one class classification approximation of SDN as shown in Eq.~\ref{eq:oneclassdecision}. SDN can parametrically model a given single shape without requiring or using large datasets. 

\vspace{-0.1in}
\section*{Acknowledgement}
We gratefully acknowledge the support of NVIDIA Corporation with the donation of the Quadro P6000 GPU used for this research.

{\small
\bibliographystyle{ieee_fullname}
\bibliography{refs}
}

\end{document}